# Examining convolutional feature extraction using Maximum Entropy (ME) and Signal-to-Noise Ratio (SNR) for image classification

Nidhi Gowdra, Roopak Sinha and Stephen MacDonell
*School of Engineering, Computer and Mathematical Sciences*
*Auckland University of Technology (AUT)*
*Auckland, New Zealand*
nidhi.gowdra@aut.ac.nz, roopak.sinha@aut.ac.nz, stephen.macdonell@aut.ac.nz

**Abstract**

*Convolutional Neural Networks (CNNs) specialize in feature extraction rather than function mapping. In doing so they form complex internal hierarchical feature representations, the complexity of which gradually increases with a corresponding increment in neural network depth. In this paper, we examine the feature extraction capabilities of CNNs using Maximum Entropy (ME) and Signal-to-Noise Ratio (SNR) to validate the idea that, CNN models should be tailored for a given task and complexity of the input data. SNR and ME measures are used as they can accurately determine in the input dataset, the relative amount of signal information to the random noise and the maximum amount of information respectively. We use two well known benchmarking datasets, MNIST and CIFAR-10 to examine the information extraction and abstraction capabilities of CNNs. Through our experiments, we examine convolutional feature extraction and abstraction capabilities in CNNs and show that the classification accuracy or performance of CNNs is greatly dependent on the amount, complexity and quality of the signal information present in the input data. Furthermore, we show the effect of information overflow and underflow on CNN classification accuracies. Our hypothesis is that the feature extraction and abstraction capabilities of convolutional layers are limited and therefore, CNN models should be tailored to the input data by using appropriately sized CNNs based on the SNR and ME measures of the input dataset.*

## 1. INTRODUCTION

CNNs are the state-of-the-art for image classification tasks verified on datasets such as MNIST [1] and CIFAR-10 [2]. The complexity of achieving these state-of-the-art performance depends on proper tuning of the HyperParameters (HPs), improper HP configurations can lead to suboptimal classification performance [3]. Some of the HPs in CNNs are learning rate, momentum, weight decay, batch-size and so on [3]. The number of tunable hyper-parameters vary with different CNN architectures, a typical CNN has around fifty unique HPs [1]. The permutations and combinations of these with respect to the number of hidden layers in any NN model amount to the order of millions.

Suboptimal configurations of HPs can lead to overfitting, where the CNN models internal representations are finely tuned to the input data causing performance degradation when new information is presented [4]. One of the reasons for overfitting is due to the fact that initialization of weights and their normalizations become too specialized for the input data [5]. In this paper, we present evidence to support the use of Signal-to-Noise Ratio (SNR) and Maximum Entropy (ME) measures to reduce the number of tunable HPs and mitigate overfitting by suggesting the use of shallower or deeper networks.

Our approach is based on the amount, complexity and quality of the signal information present in the input data measures through the use of ME and SNR values. ME and SNR measures are commonly used in digital image processing [6], [7] to precisely determine quantitative measurements for signal information present in the input data. The motivation for our novel research is based on the widespread use of SNR and ME values in image processing [8], [9], to enhance and restore degraded images. Through our experiments and analyses we show that the use of these measures to tailor CNN models yields statistically significant and improved classification performance.

## 2. BACKGROUND AND RELATED WORK

**A. Convolutional Neural Networks (CNNs)**
CNNs have been shown to outperform other types of NNs on more complex datasets like CIFAR-10,100 and ImageNet [10], [11]. CNNs are primarily feature extractors, unlike other types of DNNs which map functions. This key distinction allows CNNs to outperform other types of DNNs for nonlinear data. This is because other types of DNNs are highly susceptible to subtle variations in the input data for non-linear mapping [12]. In other words, the resiliency of CNNs arises from the fact that they extract the most prominent feature and build abstraction from the extracted information, increasing in complexity over the depth of the network.

This notion of building abstraction leading to enhanced performance is supported by authors in [13]. Forced abstraction in CNNs is based on their limited connectivity and shared weights architecture. Therefore, increasing the depth of the network also increases the extraction and generation of more complex specialized feature maps for the given input data [10].

These specialized feature maps only contain the highest variance of features which offer the greatest increases in accuracy, neglecting any subtle changes. Otherwise known as dimensionality reduction of features [2], this is a key implicit advantage of CNNs over other types of DNNs. This is because a NN that maps a function needs to account for all the features, or use a method like k-means to neglect some features. These methods are also not so versatile and fail to accurately account for dealing with outliers in the data [14].

According to several authors [10], [13], [15][16][17][18]–[19], using a CNN approach is only one of many that can be taken for accurate classification. As CNNs are computationally expensive [20], for less demanding tasks like linear regression or clustering, other types of DNNs are much more efficient [21], [22]. This is especially true for tasks that require little domain knowledge. In these instances using CNNs might actually be detrimental in terms of accuracies. Furthermore, due to the inherent computationally intense requirements of CNNs, training very large deep CNNs is impractical and leads to significant overfitting [4]. As authors in [5], [23] point out, training deep CNNs is time consuming and difficult.

**B. Maximum Entropy in Image Data**

Entropy measures are widely used in image processing for image enhancements such as de-noising and image restoration/reconstruction using de-convolutions [6], [8], [9]. Hartley Entropy (HART) or Maximum Entropy (ME) is the de-facto standard for measuring the maximum amount of information in applications of digital image processing. According to Ralph Hartley [24], the method of calculating maximum entropy is given in equation 1

$$ME = \log_2(s^n) \text{ bits} \quad (1)$$

Where, $n$ is the number of independent choices that can be made with $s$ number of distinct symbols. In grayscale images, $n$ would be 256 for the 0-255 gray levels with a 0 value for black and 255 for white and $s$ would be 784 for an image size of 28×28. ME measures are calculated separately for each of the color channels i.e. in case of color images Red, Green and Blue (RGB) and then averaged to get the final ME measure.

As an image is dependent on neighboring pixels to represent information the relative probabilities of each individual pixel are near impossible to calculate, the open-source scikit-image processing library written in python can be used to calculate the ME measures for color and grayscale images. SciKit-image processing library uses a disk (set to the size of the input training image) to scan across the input data and return the frequency count of color levels. Using this method we obtain the ME measures for MNIST and CIFAR-10 as 3.139 and 6.612 bits per pixel respectively.

**C. Signal and Noise In Image Data**

Accurate quantifiable estimation of image quality regardless of different viewing conditions plays a pivotal role in applications of digital image processing. There are many measures to mathematically calculate digital image quality, like Mean Square Error (MSE), Root Mean Square Error (RMSE), Signal-to-Noise (SNR) and Peak Signal-to-Noise (PSNR) [25]. According to the authors in [25], measures that consider the Human Visual System (HVS) that integrate perceptual quality measures offer no distinct advantages over existing methods like PSNR. As the images in MNIST and CIFAR-10 datasets are independent to one another and are randomly shuffled before training CNNs, PSNR, MSE or RMSE values cannot be calculated. Therefore, we will adopt SNR calculations to measure image quality.

Precise measurement of SNR is critical in applications of image processing, as the image might be degraded due to random noise. According to authors in [7], SNR is a measure that compares the level of desired signal to the level of background noise in the fields of science and Engineering. Mathematically, SNR in digital images is defined as the ratio of quotient of mean signal intensity to the standard deviation of the noise [26] and is given by equation 2.

$$SNR = \mu(\bar{S})/\sigma_N \quad (2)$$

Where, SNR is the signal-to-noise ratio (unit-less), $\mu(\bar{S})$ is the mean of signal data and $\sigma_N$ is the standard deviation of the signal data with respect to the random noise.

The equation for calculating the mean of signal data given in equation 3,

$$\mu(\bar{S}) = (\Sigma_{i=1}^n S)/n \quad \forall S \in (0-255). \quad (3)$$

Where, $\mu(\bar{S})$ is the mean of signal data, when the pixel values for $S$ is in between 0-255 for MNIST and 0-255 for red, green and blue color channels for CIFAR-10.

The equation for calculating standard deviation of signal data with respect to the noise is given by,

$$\sigma_N = \sqrt{(1/n)\Sigma_{i=1}^n (x_i - \mu(\bar{S})^2)} \forall S \in (0-255). \quad (4)$$

Where, $\sigma_N$ is the standard deviation of the data, i.e. the signal data with respect to the noise, $\mu(\bar{S})$ is the mean of signal data calculated using equation 3, n is the total number of pixels in the data, $x_i$ is the value of $i^{th}$ pixel in the image.

Using equation 2, we can calculate SNR for MNIST and CIFAR-10 with the mean of signal data calculated using the equation 3 and standard deviation computed using equation 4. Therefore, the SNR values for MNIST and CIFAR-10 are 0.44 and 2.40 respectively.

## 3. RELATION BETWEEN ENTROPY, SNR AND CNNS

As the theory behind CNNs suggest, the first convolution layer is able to only extract a certain quantity of signal information and every successive convolution layer gradually builds complex abstractions over this initially extracted signal information. Ideally, enough hidden layers and perfect information extraction capability should output the exact same image as the input built through abstractions achieving perfect classification accuracy. However in practice, due to random noise and other non-ideal characteristics, the classification performance of CNNs can be substantially affected.

The feature extraction capabilities of CNNs depend on the amount of information in conjunction with the quality and complexity of signal information present in the images. A low SNR indicates that the signal information is greatly corrupted by the random noise and low ME measures imply that there is lesser amount of useful signal information in the images. Images that have a higher ME values contain larger amounts of information, thus requiring a broader network, whereas images with higher SNR scores indicate that the signal information is of a higher quality and therefore allows the facilitation of deeper networks.

In circumstances where images have low SNRs like those present in the MNIST dataset, any attempts made to recover the original signal information using inverse filtering and other such methods produce outputs of unacceptable quality. This is because, according to authors in [27], noise and signal are intertwined, implying that noise in the data introduces distortions and errors which leads to uncertainties. Thus any conventional methods to extract or enhance signal information causes loss in classification accuracy by NNs, as these methods amplify the noise characteristic without significantly increasing the quality or quantity of signal. In Section 4 and, we put forth two arguments that tend to explain this non-ideal performance of CNNs using ME and SNR. In Section 4-D of this paper, we present our exploratory findings for different NN architectures and their corresponding classification accuracies with varying neural configurations using CNNs.

## 4. UNDERSTANDING INFORMATIONAL EXTRACTION AND ABSTRACTION CAPABILITIES OF CNNS

### A. Information Overflow
Information overflow is the phenomenon when there is a greater amount of signal information or a high quality of signal information in the input data than the convolution layer is able to extract or abstract. This discrepancy of greater signal information can be identified with larger ME values and quality with higher SNR values. Information overflow will occur when a single convolution layer has extracted the maximum amount of signal information possible and also if there are insufficient number of convolutional layers to completely abstract the signal information.

As an example, according to authors in [11], the CIFAR-10 dataset requires a higher number of convolutional layers when compared to the MNIST dataset. This notion is also intuitive as explained in Section 4-C1, the MNIST dataset has lower SNR and ME values when compared to CIFAR-10, thus it requires a lower number of convolution layers (depth) and fewer neurons (breadth) in each layer. If the same neural configuration (breadth and depth) was used for both datasets, it would lead to convolutional saturation causing information overflow in CIFAR-10, as it has higher SNR and ME values.

This hypothesis of information overflow can be due to two reasons, either there is a limit to the extraction capabilities of CNNs or there is a limit to the abstraction capabilities of CNNs. We test and validate these scenarios through our experimentation, the raw results are presented in Section 4-D.

### B. Information underflow
Information underflow is a phenomenon that occurs when there is a low amount and quality of signal information when compared to noise in the input data to allow for sufficient extraction and abstraction in CNNs. This phenomenon can be identified by lower ME and SNR values. Information underflow is relatively less severe in terms of afflicting the classification accuracies. Information underflow can be char-acterized by overfitting, while there are methods proposed by authors in [4], [28], [29] and to mitigate this problem by using dropout, stochastic gradient descent and hyper-parameter tuning, it exists to a certain degree in all NNs. We also propose a few techniques based on this understanding of information underflow and overflow to mitigate the problem of overfitting in Section 5.

### C. Experimentation
The raw results of experimentation are presented in Section 4-D. The tests were conducted with an intention to understand the effect of different neural configurations in the fully connected layers and varying the number and size of convolutional layers on the classification performance of the datasets with respect to information underflow and overflow. All experiments were conducted using the TFLearn front-end and a Tensorflow back-end written in python on a single 2080ti and Tesla P100 GPU with 12GB and 16GB of VRAM generously provided by InfuseAI Limited and New Zealand eScience Infrastructure respectively. The datasets used have been mentioned throughout this paper and more comprehensively in 4-C1.

We used a variant of the Visual-Geometry-Group (VGG) based CNN architecture, similar to the one proposed by authors in [23]. The variations included smaller convolutional kernels, shorter kernel strides, a dropout layer, varying arrangement of neural configurations and constant custom hyper-parameter settings. This is because as authors in [20] assert, VGG based CNNs have shown state-of-the-art performance for image recognition/classification tasks, and using dropout has shown to significantly reduce overfitting [28]. No pre-processing of the data was performed. Training-validation split was established to be 50,000-10,000 images for MNIST and 40,000-10,000 images for CIFAR-10. Experiments were performed for a limited amount of convolutional layers in-line with the initial AlexNet CNN model [11] which has only five convolutional layers [2].

**1) Datasets:** The MNIST dataset [1], consists of ten classes of 28×28 dimensional grayscale handwritten digits, differentiated by the digits in each class. The images present in CIFAR-10 dataset [2] also consist of ten classes of 32×32 dimensional tri-color (RGB) natural images, differentiated by the type of object in the images, for example cars, cats, dogs. The pixel values of both the datasets range from 0-255 for one channel in MNIST and 0-255 for three separate channels (RGB) for CIFAR-10.

### D. Results

All experiments were repeated three times and the test-set classification accuracy was averaged. The mean results are presented in Table I.

## 5. DISCUSSION

The results from Section 4-D illustrate how different neural configurations affect the accuracy of image classification. Analyzing the results, we can determine that there exists a correlation between SNR and ME values of the datasets to the classification accuracy when different breadths, depths and neural configurations were used in CNNs. We can clearly see in Table I for the MNIST dataset that increasing the breadth of convolutional layers or depth of the network by adding additional fully connected layer/s does not equate to a statistically significant difference in the classification performance of CNNs. The statistical data analyses are explained more thoroughly in Section 5-A.

The MNIST findings are in a stark contrast to the results obtained from CIFAR-10 experimentation. Experiments performed on CIFAR-10 showed a clear statistical significance when varying the breadth and depth of the network, statistical analyses is explained more thoroughly in Section 5-A. This significance can be explained by the phenomenon explained earlier in Section 4. MNIST has simple images and leads to the phenomenon of information underflow. Therefore, any changes in the neural configuration, breadth or depth of the network has minimal impact on CNN performance. Contrarily, CIFAR-10 consists of more complex images and accordingly the phenomenon of information overflow is more predominant. Therefore, any changes to the network be it variations in neural configuration, breadth or depth, affects the networks informational extraction and abstraction capabilities.

Investigating data from Table I, we can claim that CNNs features extraction and abstraction capabilities correspond to the phenomena of information underflow and overflow. The attributes for these phenomena can be quantified using ME and SNR values. The higher the ME and SNR values, the greater will be the amount and quality of signal information in the datasets respectively. We can now postulate based on our experimental results that any changes to the neural configurations, breadth, depth and HPs of the CNN model will affect the informational extraction and abstraction capabilities. Furthermore, information underflow leads to the network overfitting on the noise rather than the signal information. Information overflow leads to the network being unable to extract enough signal information for further abstraction. Both these cases are detrimental for CNN performance.

TABLE I. TABLE OF RESULTS COMPARING THE TEST-SET CLASSIFICATION ACCURACIES FOR MNIST AND CIFAR-10 DATASETS

| Configuration | Epochs | Acc. in % |
|---|---|---|
| **MNIST Dataset** | | |
| 32 Convolutional kernels and 784 neuron FC layer | 10, 500, 1,000, 1,500, 2,000 | 93.3, 97.0, 97.0, 97.1, 97.2 |
| 32-64 Convolutional kernels and 784 neuron FC layer | 10, 500, 1,000, 1,500, 2,000 | 94.5, 96, 98.5, 97.7, 96.5 |
| 32 Convolutional kernels and 784-392 neuron FC layers | 10, 500, 1,000, 1,500, 2,000 | 94.2, 96.6, 96.3, 96.89, 97.3 |
| 32-64 Convolutional kernels and 784-392 neuron FC layers | 10, 500, 1,000, 1,500, 2,000 | 95.4, 97.4, 98.2, 98.4, 98.9 |
| 32 Convolutional kernels and 784-512 neuron FC layers | 10, 500, 1,000, 1,500, 2,000 | 93.5, 97.5, 97.39, 97.2, 98.1 |
| 32-64 Convolutional kernels and 784-512 neuron FC layers | 10, 500, 1,000, 1,500, 2,000 | 93.89, 97.5, 97.5, 98.1, 97.9 |
| 32 Convolutional kernels and 784-784 neuron FC layers | 10, 500, 1,000, 1,500, 2,000 | 95.19, 96.6, 96.7, 97.5, 97.39 |
| 32-64 Convolutional kernels (2 layers) and 784-784 neuron FC layers | 10, 500, 1,000, 1,500, 2,000 | 94.19, 96.0, 98.0, 98.4, 97.6 |
| **CIFAR-10 Dataset** | | |
| 32 Convolutional kernels and 3072 neuron FC layer | 10, 500, 1,000, 1,500, 2,000 | 21.4, 10.9, 10.2, 10.1, 10.0 |
| 32-64 Convolutional kernels and 3072-3072 neuron FC layers | 10, 500, 1,000, 1,500, 2,000 | 43.5, 52.4, 57.12, 55.19, 53.35 |
| 32-64 Convolutional kernels and 3072-1536 neuron FC layers | 10, 500, 1,000, 1,500, 2,000 | 45.5, 56.64, 59.43, 61.30, 60.82 |
| 32-64-128 Convolutional kernels and 3072-1024-512 neuron FC layers | 10, 500, 1,000, 1,500, 2,000 | 43.5, 51.0, 55.2, 58.5, 61.5 |
| 32-64-128 Convolutional kernels and 3072-1536-768 neuron FC layers | 10, 500, 1,000, 1,500, 2,000 | 60.1, 67.2, 67.45, 68.58, 68.49 |
| 32-64-128 Convolutional kernels and 3072-3072-3072 neuron FC layers | 10, 500, 1,000, 1,500, 2,000 | 44.1, 53.8, 55.2, 60.0, 53.4 |

*Based on these observations, we recommend adopting a broader convolutional and FC layers with halving the number of units for each successive layer/s added until the phenomenon of underflow is observed. the depth of the network is restricted to the initial breadth and thus might become a limiting factor for more complex datasets due to the associated memory and computational constraints.*

### A. Statistical Significance

To ensure reproducibility and to eliminate any differences in the data due to pure random chance, all our experimental results are averaged across three separate instances. We employed two-tailed paired t-tests with the standard scientific threshold of 95% or p-values below 0.05, to accurately highlight statistically significant differences in the data. A two-tailed t-test was employed as there is no prior knowledge or estimate to determine if the results would be positive or negative. This rigorous methodology also ensures any interpretations drawn from the experimental data to be empirically valid. Our null hypothesis during testing being that there would be no statistical significance.

**1) Experiments on the MNIST dataset:** To analyze any significant differences in the neural configurations we ran t-tests on the averaged results presented in Table I for the MNIST dataset. The first test was to examine if there were any differences between increasing the convolutional layers while keeping the fully connected layers constant at one layer (1 neuron for every pixel). There was no observed difference with a p-value of 0.558. The next test was to examine if there were any differences between the neural configurations keeping the convolutional layers constant. There was no observable differences with p-values of 0.0816, 0.2065, 0.9046 for the combinations of one convolutional layer (32 convolutional kernels). There was no observable difference for the combinations of two convolutional layer with p-values of 0.0508, 0.7212, 0.0508.

The third test was to examine the efficacy of convolutional layers while keeping the Fully Connected (FC) layers constant at 784-392 neurons for the first and second FC layer respectively. There was a statistical difference when increasing the number of convolutional layers by doubling the number of filters in each layer and decreasing the number of neurons in each fully connected layer by half, with a p-value of 0.0018. In other words between one convolutional layer (32 convolutional kernels) and two FC layers with 784 and 392 neurons; two convolutional layers (32 and 64 convolutional kernels) and two FC layers with 784 and 392 neurons. These results indicate that abstraction needed to be forced in cases of information underflow and different neural configuration have little effect on the performance of CNNs.

**2) Experiments on the Cifar-10 dataset:** Performing similar tests on the CIFAR-10 dataset, we can clearly see statistically significant differences. The effect of increasing convolutional layers and adopting different variations in neural configuration had a great impact on the overall classification accuracies of the test-set. Performing a t-test on two convolutional layers with 32 and 64 convolutional kernels and two FC layers with 3072 and 1536 neurons; three convolutional layers with 32,64 and 128 convolutional kernels and three FC layers with 3072,3072 and 3072 neurons, we observe a statistical significance with a p-value of 0.0382. Performing a t-test on two convolutional layers with 32 and 64 convolutional kernels and two FC layers with 3072 and 3072 neurons; two convolutional layers with 32 and 64 convolutional kernels and two fully connected with 3072 and 1536 neurons yields a statistical significance with a p-value of 0.0142.

These same significant differences appear in results for three convolutional configurations with p-values of 0.0027 and 0.0006 when using fully connected layers with halved number of neurons per each successive layer. There was no difference between diverse neural configurations. That is, between three convolutional (32-64-128 filter per layer), three fully connected layer (3072-3072-3072 neurons per layer) and three convolutional (32-64-128 filter per layer), three fully connected layer (3072-1024-512 neurons per layer) yielding a p-value of 0.7561.

We can infer that neural configurations play an important role where information overflow occurs with the best results obtained when the neurons are halved for every successive layer with the initial layer having one neuron for every pixel of the input image. We can further postulate that in order to achieve improved classification performance there should be forced abstraction in the fully connected layers proportional to that of the convolutional layer abstraction. It would suggest that these abstractions need to be encoded and decoded along the full depth of the CNN model, which is in-line with the research conducted by multiple researchers [30][31]–[32].

# 6. CONCLUSION

In this paper, in Section 1, we looked at some of the inherent problems arising in CNNs and explained the research problem. In Section 2, we thoroughly examined existing literature to explain the working of DNNs, CNNs along with their advantages and drawbacks. We also briefly explained methods to quantify the amount (ME) and quality (SNR) of images in datasets. In Section 3, we explored the relationship between entropy, SNR and CNNs. In Section 4, we discussed the novel application of information theory to the informational extraction and abstraction capabilities of CNNs using entropy and SNR measures. In Section 4-D, we presented our experimental findings in Table I for the MNSIT and CIFAR-10 datasets by varying neural configurations, breadth and depth of the CNN. In Section 5, we discussed our results and explained the effect of information underflow and overflow phenomena on performance of CNNs. In Section 6-A, we identified the limitations of our experimentation and proposed claims.

To summarize, the claims we make in this paper are, CNNs experience the phenomenon of information underflow when there is a relatively low amount of total signal information including complexity. In this instance the quality of signal is also highly corrupted by random noise, therefore the CNN will suffer in performance when deeper networks are used. The phenomenon of information overflow will arise when there is a higher amount of signal information and is not relatively corrupted by random noise. These attributes of the data can be measured using the metrics of ME and SNR.

Our reasoning behind these claims are based on the theory that a single convolution operation isolates only one feature while suppressing the rest. Feature abstraction is built using the previous simpler features to form complex representations. Furthermore, our recommendation that using shallower but broader networks for simplistic datasets and deeper but narrower networks for more complex datasets is in close agreement with the experimental findings presented by authors in [11].

In conclusion, we present experimental evidence to validate our claim that CNNs have limits on their informational feature extraction and abstraction capabilities. The data also compels us to claim that varying neural configuration, breadth, depth and other hyper-parameter settings have minimal impact on simplistic datasets, but have statistically significant repercussions on more complex datasets. Using the metrics of ME and SNR we can show that there exists a clear relation to optimally tune CNN architectures tailored for the input data.

## A. Limitations

The key limitation of this study is that the relationship between ME, SNR and information extraction capabilities of neural networks along with the phenomena of information overflow and underflow are tested on CNNs with only two datasets. We do maintain that these same phenomena and general extraction, abstraction capabilities should apply for other types of DNNs, but this claim has not been empirically tested. Another limitation that can be identified is that our claim has been experimentally validated against spatially variant data, i.e. temporal aspects have not been considered for this paper. Finally, the calculated ME and SNR measures for the multi-class datasets used were averaged across all the ten classes and further investigation into the information underflow and overflow phenomenon for individual classes is an interesting area of research that will be explored along with the other limitations in future publications.

# REFERENCES

[1] Y. LeCun, L. Bottou, Y. Bengio, and P. Haffner, "Gradient-based learning applied to document recognition," Proceedings of the IEEE, vol. 86, no. 11, pp. 2278–2324, 1998.
[2] A. Krizhevsky and G. Hinton, "Learning multiple layers of features from tiny images," 2009.
[3] A. Coates, A. Ng, and H. Lee, "An analysis of single-layer networks in unsupervised feature learning," in Proceedings of the fourteenth international conference on artificial intelligence and statistics, 2011, pp. 215–223.
[4] D. M. Hawkins, "The problem of overfitting," Journal of chemical information and computer sciences, vol. 44, no. 1, pp. 1–12, 2004.
[5] X. Glorot and Y. Bengio, "Understanding the difficulty of training deep feedforward neural networks," in Proceedings of the thirteenth international conference on artificial intelligence and statistics, 2010, pp. 249–256.
[6] S. F. Gull and J. Skilling, "Maximum entropy method in image processing," in IEE Proceedings F (Communications, Radar and Signal Processing), vol. 131(6). IET, 1984, pp. 646–659.
[7] R. C. Gonzalez and R. E. Woods, "Image processing," Digital image processing, vol. 2, 2007.
[8] Z. Krbcova and J. Kukal, "Relationship between entropy and snr changes in image enhancement," EURASIP Journal on Image and Video Processing, vol. 2017, no. 1, p. 83, 2017.
[9] M.-A. Petrovici, C. Damian, and D. Coltuc, "Maximum entropy principle in image restoration," ADVANCES IN ELECTRICAL AND COM- PUTER ENGINEERING, vol. 18, no. 2, pp. 77–84, 2018.
[10] A. Krizhevsky and G. Hinton, "Convolutional deep belief networks on cifar-10," Unpublished manuscript, vol. 40, 2010.
[11] A. Krizhevsky, I. Sutskever, and G. E. Hinton, "Imagenet classification with deep convolutional neural networks," in Advances in neural information processing systems, 2012, pp. 1097–1105.
[12] G. Bebis and M. Georgiopoulos, "Feed-forward neural networks," IEEE Potentials, vol. 13, no. 4, pp. 27–31, 1994.
[13] Y. LeCun, F. J. Huang, and L. Bottou, "Learning methods for generic object recognition with invariance to pose and lighting," in Computer Vision and Pattern Recognition, 2004. CVPR 2004. Proceedings of the 2004 IEEE Computer Society Conference on, vol. 2. IEEE, 2004, pp. II–104.
[14] J. A. Hartigan and M. A. Wong, "Algorithm as 136: A k-means clustering algorithm," Journal of the Royal Statistical Society. Series C (Applied Statistics), vol. 28, no. 1, pp. 100–108, 1979.
[15] K. Jarrett, K. Kavukcuoglu, Y. LeCun et al., "What is the best multi-stage architecture for object recognition?" in Computer Vision, 2009 IEEE 12th International Conference on. IEEE, 2009, pp. 2146–2153.
[16] H. Lee, R. Grosse, R. Ranganath, and A. Y. Ng, "Convolutional deep belief networks for scalable unsupervised learning of hierarchical representations," in Proceedings of the 26th annual international conference on machine learning. ACM, 2009, pp. 609–616.
[17] Y. LeCun, B. E. Boser, J. S. Denker, D. Henderson, R. E. Howard, W. E. Hubbard, and L. D. Jackel, "Handwritten digit recognition with a back-propagation network," in Advances in neural information processing systems, 1990, pp. 396–404.
[18] N. Pinto, D. Doukhan, J. J. Di Carlo, and D. D. Cox, "Ahigh-throughput screening approach to discovering good forms of biologically inspired visual representation," PLoS computational biology, vol. 5, no. 11, p. e1000579, 2009.
[19] S. C. Turaga, J. F. Murray, V. Jain, F. Roth, M. Helmstaedter, K. Briggman, W. Denk, and H. S. Seung, "Convolutional networks can learn to generate affinity graphs for image segmentation," Neural computation, vol. 22, no. 2, pp. 511–538, 2010.
[20] A. Ardakani, C. Condo, M. Ahmadi, and W. J. Gross, "An architecture to accelerate convolution in deep neural networks," IEEE Transactions on Circuits and Systems I: Regular Papers, vol. 65, no. 4, pp. 1349–1362, 2018.
[21] Y. LeCun, Y. Bengio, and G. Hinton, "Deep learning," Nature, vol. 521, no. 7553, pp. 436–444, 2015.
[22] D. J. Becker, T. Sterling, D. Savarese, J. E. Dorband, U. A. Ranawak, and C. V. Packer, "Beowulf: A parallel workstation for scientific computation," in Proceedings, International Conference on Parallel Processing, vol. 95, 1995, pp. 11–14.
[23] K. Simonyan and A. Zisserman, "Very deep convolutional networks for large-scale image recognition," arXiv preprint arXiv:1409.1556, 2014.
[24] R. V. Hartley, "Transmission of information," Bell Labs Technical Journal, vol. 7, no. 3, pp. 535–563, 1928.
[25] Z. Wang and A. C. Bovik, "A universal image quality index," IEEE signal processing letters, vol. 9, no. 3, pp. 81–84, 2002.
[26] M. Welvaert and Y. Rosseel, "On the definition of signal-to-noise ratio and contrast-to-noise ratio for fmri data," PLoS one, vol. 8, no. 11, p. e77089, 2013.
[27] J. R. Pierce, "An introduction to information theory. symbols, signals and noise., revised edition of symbols, signals and noise: the nature and process of communication (1961)," 1980.
[28] N. Srivastava, G. Hinton, A. Krizhevsky, I. Sutskever, and R. Salakhutdinov, "Dropout: A simple way to prevent neural networks from over- fitting," The Journal of Machine Learning Research, vol. 15, no. 1, pp. 1929–1958, 2014.
[29] R. Caruana, S. Lawrence, and C. L. Giles, "Overfitting in neural nets: Backpropagation, conjugate gradient, and early stopping," in Advances in neural information processing systems, 2001, pp. 402–408.
[30] G. E. Hinton and R. R. Salakhutdinov, "Reducing the dimensionality of data with neural networks," science, vol. 313, no. 5786, pp. 504–507, 2006.
[31] V. Badrinarayanan, A. Kendall, and R. Cipolla, "Segnet: A deep convolutional encoder-decoder architecture for image segmentation," arXiv preprint arXiv:1511.00561, 2015.
[32] I. Sutskever, O. Vinyals, and Q. V. Le, "Sequence to sequence learning with neural networks," in Advances in neural information processing systems, 2014, pp. 3104–3112.